\documentclass[10pt,twocolumn,letterpaper]{article}

\usepackage{iccv}
\usepackage{times}
\usepackage{epsfig}
\usepackage{graphicx}
\usepackage{amsmath}
\usepackage{amssymb}

\usepackage{multirow}
\usepackage[pagebackref=true,breaklinks=true,letterpaper=true,colorlinks,bookmarks=false]{hyperref}

\def\ie{\emph{i.e.}}
\def\eg{\emph{e.g.}}
\def\etal{{\em et al.}}
\def\etc{{\em etc.}}

\iccvfinalcopy 

\def\iccvPaperID{3762} 

\ificcvfinal\fi
\begin{document}

\title{Deep Floor Plan Recognition Using a Multi-Task Network\\ with Room-Boundary-Guided Attention}

\author{Zhiliang Zeng\quad Xianzhi Li\quad Ying Kin Yu\quad Chi-Wing Fu\\
	The Chinese University of Hong Kong \\ 
	{\tt\small \{zlzeng,xzli,cwfu\}@cse.cuhk.edu.hk}\hspace{10mm}{\tt\small ykyu.hk@gmail.com}\qquad
}

\maketitle

\begin{abstract}
This paper presents a new approach to recognize elements in floor plan layouts. Besides walls and rooms, we aim to recognize diverse floor plan elements, such as doors, windows and different types of rooms, in the floor layouts. 
To this end, we model a hierarchy of floor plan elements and design a deep multi-task neural network with two tasks: one to learn to predict room-boundary elements, and the other to predict rooms with types. More importantly, we formulate the room-boundary-guided attention mechanism in our spatial contextual module to carefully take room-boundary features into account to enhance the room-type predictions. Furthermore, we design a cross-and-within-task weighted loss to balance the multi-label tasks and prepare two new datasets for floor plan recognition. Experimental results demonstrate the superiority and effectiveness of our network over the state-of-the-art methods. 
%
%
\end{abstract}

\section{Introduction}

%
To recognize floor plan elements in a layout requires the learning of semantic information in the floor plans.
It is not merely a general segmentation problem since floor plans present not only the individual floor plan elements, such as walls, doors, windows, and closets,~\etc, but also how the elements relate to one another, and how they are arranged to make up different types of rooms.
While recognizing semantic information in floor plans is generally straightforward for humans, automatically processing floor plans and recognizing layout semantics is a very challenging problem in image understanding and document analysis.

\begin{figure}[!t]
\centering
\includegraphics[width=0.99\linewidth]{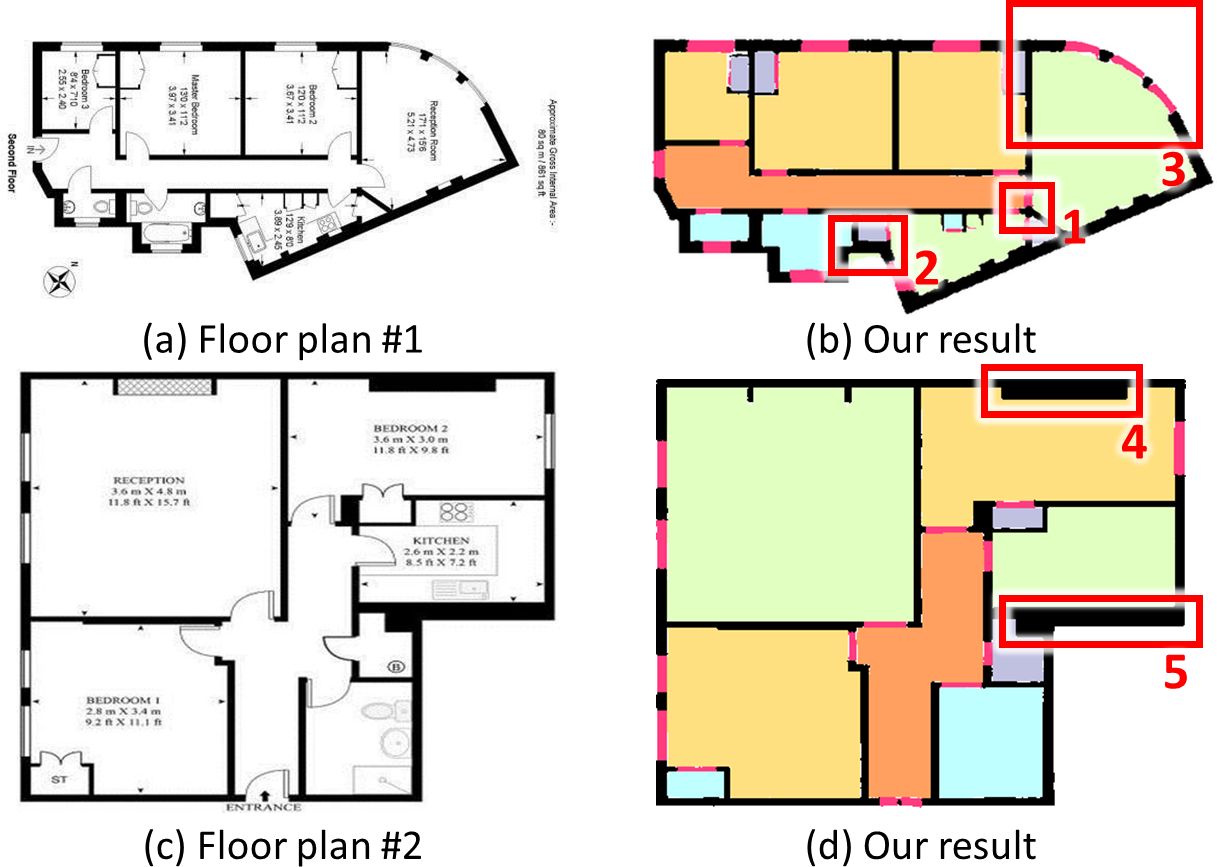}
\vspace*{0.5mm}
\caption{Our network is able to recognize walls of nonuniform thickness (see boxes 2, 4, 5), walls that meet at irregular junctions (see boxes 1, 2), curved walls (see box 3), and various room types in the layout; see Figure~\ref{fig3_1} for the legend of the color labels.}
\label{fig1}
\vspace*{-2.5mm}
\end{figure}

Traditionally, the problem is solved based on low-level image processing methods~\cite{mace2010system,ahmed2011improved,gimenez2016automatic} that exploit heuristics to locate the graphical notations in the floor plans.
Clearly, simply relying on hand-crafted features is insufficient, since it lacks generality to handle diverse conditions.

Recent methods~\cite{Raster2Vector,dodge2017parsing,yamasaki2018apartment} for the problem has begun to explore deep learning approaches.
Liu~\etal~\cite{Raster2Vector} designed a convolutional neural network (CNN) to recognize junction points in a floor plan image and connected the junctions to locate walls.
The method, however, can only locate walls of uniform thickness along XY-principal directions in the image.
%
Later, Yamasaki~\etal~\cite{yamasaki2018apartment} adopted a fully convolutional network to label pixels in a floor plan; however, the method simply uses a general segmentation network to 
recognize pixels of different classes and ignores the spatial relations between floor plan elements and room boundary.


This paper presents a new method for floor plan recognition, with a focus on recognizing diverse floor plan elements,~\eg, walls, doors, rooms, closets,~\etc; see Figure~\ref{fig1} for two example results and Figure~\ref{fig3_1} for the legend. These elements are inter-related graphical elements with structural semantics in the floor plans. To approach the problem, we model {\em a hierarchy of labels\/} for the floor plan elements and design a {\em deep multi-task neural network\/} based on the hierarchy.
Our network learns shared features from the input floor plan and refines the features to learn to recognize individual elements. Specifically, we design the {\em spatial contextual module\/} to explore the spatial relations between elements via the {\em room-boundary-guided attention mechanism\/} to avoid feature blurring, and formulate the {\em cross-and-within-task weighted loss\/} to balance the labels across and within tasks.
Hence, we can effectively explore the spatial relations between the floor plan elements to maximize the network learning; see again the example results shown in Figure~\ref{fig1}, which exhibit the capability of our network.
%

Our contributions are threefold.
%
First, we design a deep multi-task neural network to learn the spatial relations between floor plan elements to maximize network learning.
%
Second, we present the spatial contextual module with the room-boundary-guided attention mechanism to learn the spatial semantic  information, and formulate the cross-and-within-task weighted loss to balance the losses for our tasks.
%
Lastly, we take the datasets from~\cite{Raster2Vector} and~\cite{Rent3D}, collect additional floor plans, and prepare two new datasets with labels on various floor plan elements and room types.
\section{Related Work}


%

Traditional approaches recognize elements in floor plan based on low-level image processing.
Ryall~\etal~\cite{ryall1995semi} applied a semi-automatic method for room segmentation.
Other early methods~\cite{ah1997variations, dosch2000complete} locate walls, doors, and rooms by detecting graphical shapes in the layout,~\eg, line, arc, and small loop.
Or~\etal~\cite{or2005highly} converted bitmapped floor plans to vector graphics and generated 3D room models.
Ahmed~\etal~\cite{ahmed2011improved} separated text from graphics and extracted lines of various thickness, where walls are extracted from the thicker lines and symbols are assumed to have thin lines; then, they applied such information
to further locate doors and windows.
%
Gimenez~\etal~\cite{gimenez2016automatic} recognized walls and openings using heuristics, and generated 3D building models based on the detected walls and doors.

Using heuristics to recognize low-level elements in floor plans is error-prone.
This motivates the development of machine learning methods~\cite{de2011wall}, and very recently, deep learning methods~\cite{dodge2017parsing,Raster2Vector,yamasaki2018apartment} to address the problem.
Dodge~\etal~\cite{dodge2017parsing} used a fully convolutional network (FCN) to first detect the wall pixels, and then adopted a faster R-CNN framework to detect doors, sliding doors, and symbols such as kitchen stoves and bathtubs.
Also, they employed a library tool to recognize text to estimate the room size.
%
%

Liu~\etal~\cite{Raster2Vector} trained a deep neural network to first identify junction points in a given floor plan image, and then used integer programming to join the junctions to locate the walls in the floor plan.
Due to the Manhattan assumption, the method can only handle walls that align with the two principal axes in the floor plan image.
Hence, it can recognize layouts with only rectangular rooms and walls of uniform thickness.
%
%
Later, Yamasaki~\etal~\cite{yamasaki2018apartment} trained a FCN to label the pixels in a floor plan with several classes.
The classified pixels formed a graph model and were taken to retrieve houses of similar structures.
However, their method adopts a general segmentation network, where it simply recognizes pixels of different classes independently, thus ignoring the spatial relations among classes in the inference.

\begin{figure}[!t]
\centering
\includegraphics[width=0.9\linewidth]{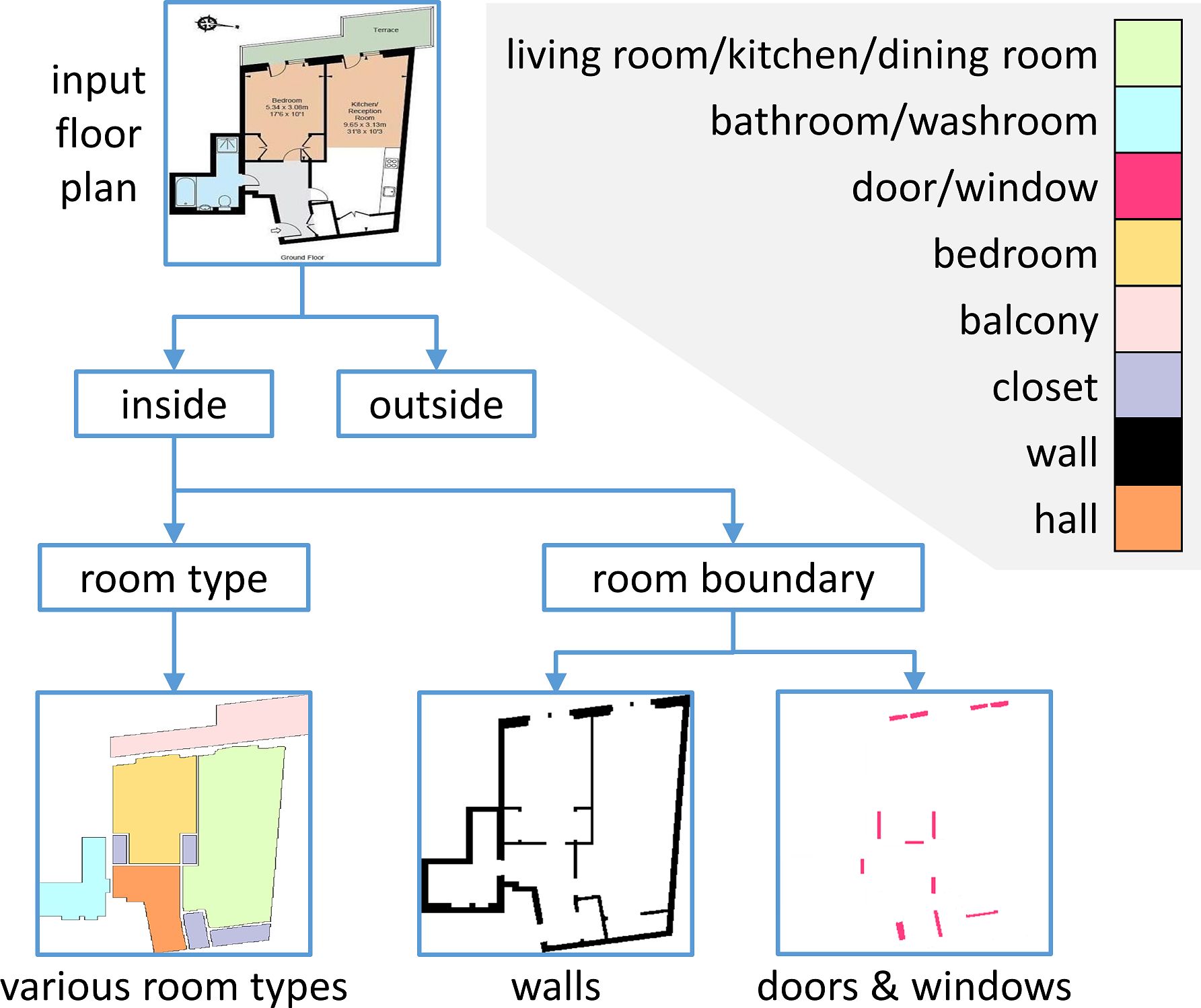}
\caption{Floor plan elements organized in a hierarchy.}
\label{fig3_1}
\vspace*{-2mm}
\end{figure}

Compared with the recent works, our method has several distinctive improvements.
Technical-wise, our method simultaneously considers multiple floor plan elements in the network; particularly, we take their spatial relationships into account and design a multi-task approach to maximize the learning of the floor plan elements in the network.
Result-wise, our method is more general and capable of recognizing nonrectangular room layouts and walls of nonuniform thickness, as well as various room types; see Figure~\ref{fig3_1}.

Recently, there are several other works 
~\cite{Zhang_2014_ECCV,Lee_2017_ICCV,Zou_2018_CVPR,Yang_2019_CVPR,Sun_2019_CVPR} related to room layouts, but they focus on a different problem,~\ie, to reconstruct 3D room layouts from photos.

\section{Our Method}

\begin{figure*}[!htb]
\centering
\includegraphics[width=0.98\linewidth]{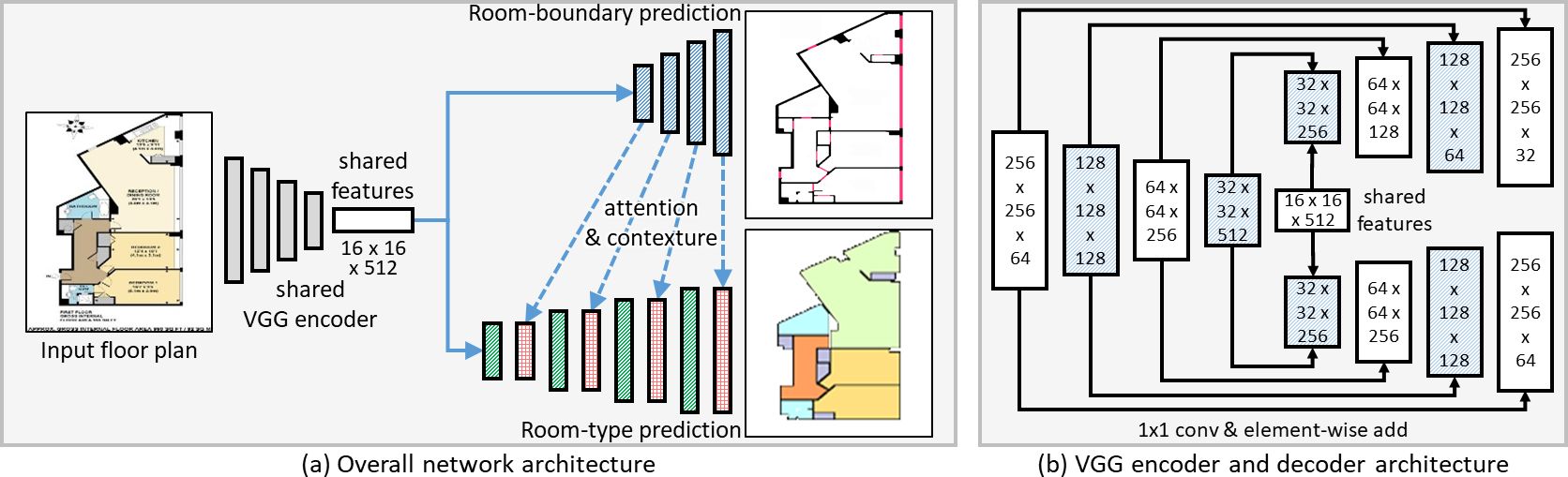}
\caption{(a) Schematic diagram illustrating our deep multi-task neural network. We have a VGG encoder to extract features from the input floor plan image.
These features are shared for two subsequent tasks in the network: one for predicting the room-boundary pixels (wall, door, and windows) and the other for predicting the room-type pixels (dining room, bedroom, etc.).
Most importantly, these two tasks have separate VGG decoders. We design the room-boundary-guided attention mechanism (blue arrows) to make use of the room-boundary features from the decoder in the upper branch to help the decoder in the lower path to learn the contextual features (red boxes) for predicting the room-type pixels.
(b) Details of the VGG encoder and decoders. The dimensions of the features in the network are shown.}
%
%
\label{fig3_2}
\vspace*{-2mm}
\end{figure*}


\subsection{Goals and Problem Formulation}


The objectives of this work are as follows.
%
First, we aim to recognize various kinds of floor plan elements, which are not only limited to walls but also include doors, windows, room regions,~\etc~
Second, we target to handle rooms of nonrectangular shapes and walls of nonuniform thickness.
Last, we aim also to recognize the rooms types in floor plans,~\eg, dining room, bedroom, bathroom,~\etc


Achieving these goals requires the ability to process the floor plans and find multiple nonoverlapping but spatially-correlated elements in the plans.
In our method, we first organize the floor plan elements in a hierarchy (see Figure~\ref{fig3_1}), where pixels in a floor plan can be identified as inside or outside, while the inside pixels can be further identified as {\em room-boundary pixels\/} or {\em room-type pixels\/}.
Moreover, the room-boundary pixels can be {\em walls\/}, {\em doors\/}, or {\em windows\/}, whereas room-type pixels can be the {\em living room, bathroom, bedroom, etc.\/}; see the legend in Figure~\ref{fig3_1}.
Based on the hierarchy, we design a deep multi-task network with one task to predict room-boundary elements and the other to predict room-type elements.
In particular, we formulate the spatial contextual module to explore the spatial relations between elements,~\ie, using the features learned for the room boundary to refine the features for learning the room types.

\subsection{Network Architecture}




\vspace*{-0.5mm}
\paragraph{Overall network architecture.}
Figure~\ref{fig3_2}(a) presents the overall network architecture.
First, we adopt a shared VGG encoder~\cite{Simonyan14c} to extract features from the input floor plan image.
Then, we have two main tasks in the network: one for predicting the room-boundary pixels with three labels,~\ie, wall, door, and window, and the other for predicting the room-type pixels with eight labels,~\ie, dining room, washroom,~\etc; see Figure~\ref{fig3_1} for details.
Here, \emph{room boundary} refers to the floor-plan elements that separate room regions in floor plans; it is not simply low-level edges nor the outermost border that separates the foreground and background.

Specifically, our network first learns the shared feature, common for both tasks, then makes use of two separate VGG decoders (see Figure~\ref{fig3_2}(b) for the connections and feature dimensions) to perform the two tasks.
Hence, the network can learn additional features for each task.
To maximize the network learning, we further make use of the room-boundary context features to bound and guide the discovery of room regions, as well as their types; here, we design the spatial contextual module to process and pass the room-boundary features from the top decoder (see Figure~\ref{fig3_2}(a)) to the bottom decoder to maximize the feature integration for room-type predictions.

\begin{figure*}[!t]
\centering
\includegraphics[width=14.5cm]{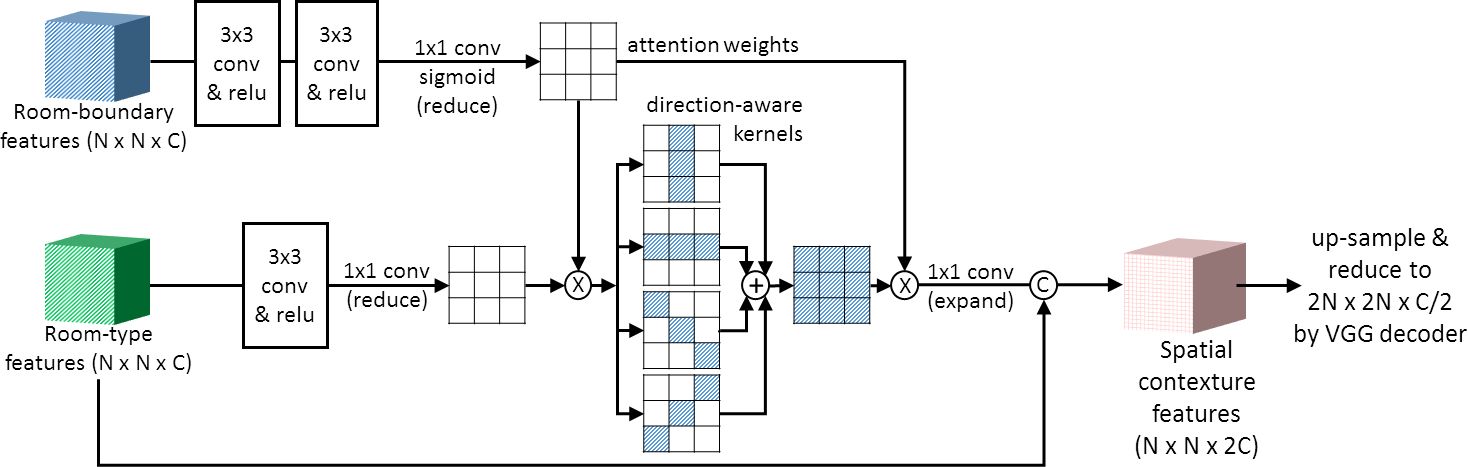}
\caption{Our {\em spatial contextual module\/} with the {\em room-boundary-guided attention mechanism\/}, which leverages the room-boundary features to learn the attention weights for room-type prediction.
In the lower branch, we use convolutional layers with four different direction-aware kernels to generate features for integration with the attention weights and produce the spatial contextual features (in red; see also Figure~\ref{fig3_2}).
Here ``C'' denotes concatenation, while ``X'' and ``+'' denote element-wise multiplication and addition, respectively.}
%
\label{fig3_4}
\vspace*{-3mm}
\end{figure*}




\vspace*{-3.5mm}
\paragraph{Spatial contextual module.}
%
Figure~\ref{fig3_4} shows the network architecture of the spatial contextual module.
It has two branches.
The input to the top branch is the room-boundary features from the top VGG decoder (see the blue boxes in Figures~\ref{fig3_2}(a) \&~\ref{fig3_4}), while the input to the bottom branch is the room-type features from the bottom VGG decoder (see the green boxes in Figures~\ref{fig3_2}(a) \&~\ref{fig3_4}).
See again Figure~\ref{fig3_2}(a): there are four levels in the VGG decoders, and the spatial contextual module (see the dashed arrows in Figure~\ref{fig3_2}(a)) is applied four times, once per level, to integrate the room-boundary and room-type features from the same level (\ie, in the same resolution) and generate the spatial contextual features; see the red boxes in Figures~\ref{fig3_2}(a) \&~\ref{fig3_4}.
\begin{itemize}
\vspace*{-1mm}
\item
In the top branch, we apply a series of convolutions to the room-boundary feature and reduce it to a 2D feature map as the {\em attention weights\/}, denoted as $a_{m,n}$ at pixel location $m,n$. The attention weights are learned through the convolutions rather than being fixed.
\vspace*{-1mm}
\item
Furthermore, we apply the attention weights to the bottom branch twice; see the ``X'' operators in Figure~\ref{fig3_4}.
The first attention is applied to compress the noisy features before the four convolutional layers with direction-aware kernels, while the second attention is applied to further suppress the blurring features.
We call it the {\em room-boundary-guided attention mechanism\/} since the attention weights are learned from the room-boundary features.
Let $f_{m,n}$ as the input feature for the first attention weight $a_{m,n}$ and $f'_{m,n}$ as the output, the X operation can be expressed as
\vspace*{-2mm}
\begin{equation}\label{eq4_1}
  f'_{m,n} = a_{m,n} \cdot f_{m,n} \ .
\end{equation}
\vspace*{-8mm}
\item
In the bottom branch as shown in Figure~\ref{fig3_4}, we first apply a $3\times3$ convolution to the room-type features and then reduce it into a 2D feature map.
After that, we apply the first attention to the 2D feature map followed by four separate direction-aware kernels (horizontal, vertical, diagonal, and flipped diagonal) of $k$ unit size to further process the feature.
Taking the horizontal kernel as an example, our equation is as follows:
\vspace*{-1mm}
\begin{equation}
\label{eq_group3}
\begin{aligned}
h_{m,n} =& \sum_k(\alpha_{m-k,n} \cdot f'_{m-k,n} + \alpha_{m,n} \cdot f'_{m,n} \\&
+ \alpha_{m+k,n} \cdot f'_{m+k,n}),
\end{aligned}
\end{equation}
where $h_{m,n}$ is the contextual features along the horizontal direction;
$f'_{m,n}$ is the input feature (see Eq.~\eqref{eq4_1}); and
$\alpha$ is the weight.
In our experiments, we set $\alpha$ to $1$.
\vspace*{-4mm}
\item
In the second attention, we further apply the attention weights ($a_{m,n}$) to integrate the aggregated features:
\vspace*{-1mm}
\begin{equation}\label{eq4}
f''_{m,n} = a_{m,n} \cdot (h_{m,n} + v_{m,n} + d_{m,n} + d'_{m,n}),
\end{equation}
where $v_{m,n}$, $d_{m,n}$, and $d'_{m,n}$ denotes the contextual features along the vertical, diagonal, and flipped diagonal directions, respectively, after the convolutions with the direction-aware kernels.
\end{itemize}

\begin{figure*}[t]
	\centering
	\includegraphics[width=0.98\linewidth]{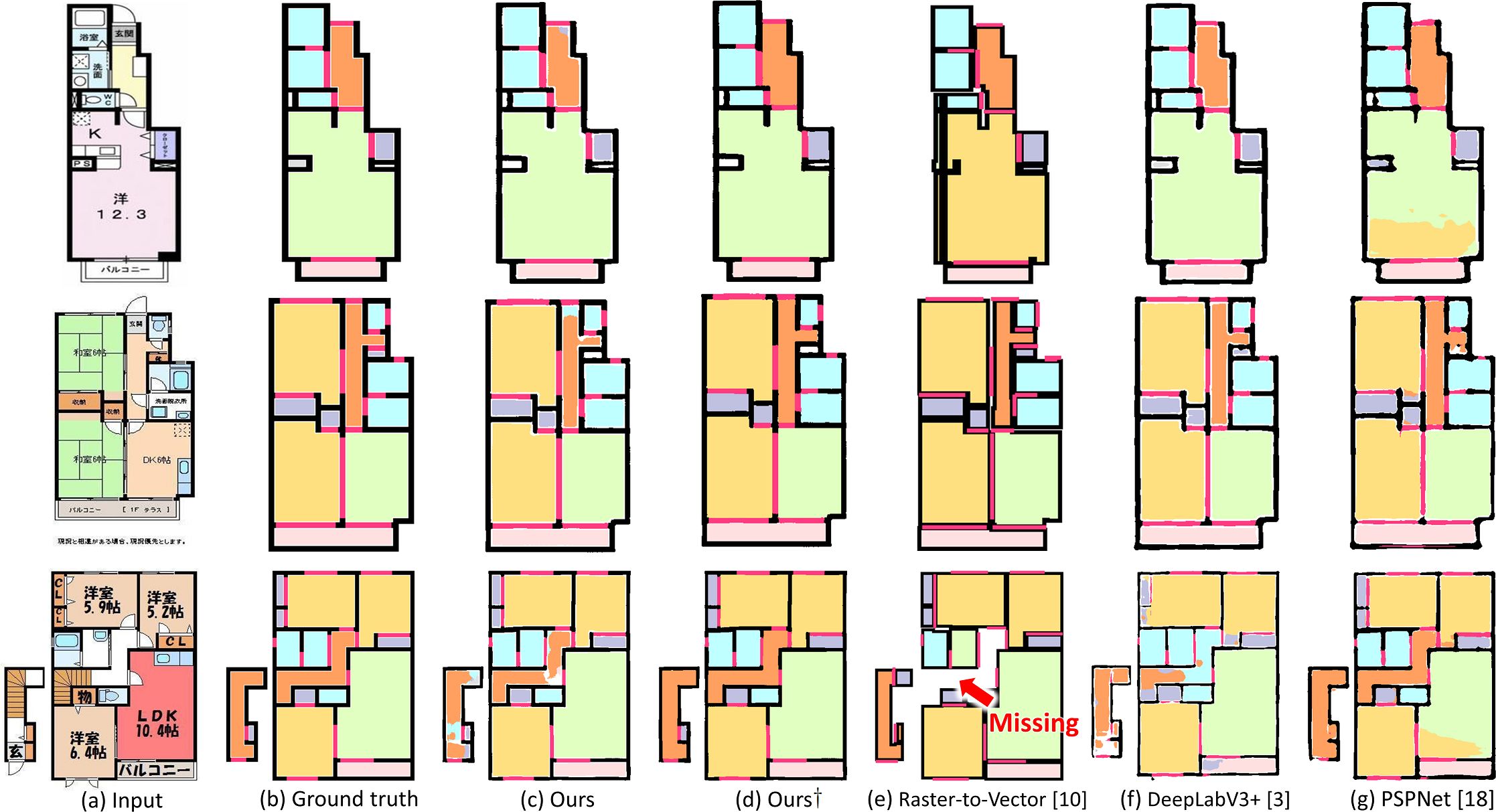}
	\caption{Visual comparison of floor plan recognition results produced by our method (c\&d) and by others (e-g) on the R2V dataset; note that we have to use rectangular floor plans for comparison with Raster-to-Vector~\cite{Raster2Vector}.
Symbol $\dag$ indicates the postprocessing step.}
	\label{fig4_1}
\end{figure*}

\begin{table*}[t]
	\centering
	\caption{Comparison with Raster-to-Vector~\cite{Raster2Vector} on the R2V dataset.
Symbol $\dag$ indicates our method with postprocessing (see Section~\ref{ssec:implem}).}
	\vspace*{1mm}
	\label{table4_5}
	\resizebox{0.9\linewidth}{!}{
		\begin{tabular}{@{\hspace{1mm}}c@{\hspace{1mm}}||@{\hspace{1mm}}c@{\hspace{1mm}}|@{\hspace{1mm}}c@{\hspace{1mm}}|@{\hspace{1mm}}c@{\hspace{1mm}}|@{\hspace{1mm}}c@{\hspace{1mm}}|@{\hspace{1mm}}c@{\hspace{1mm}}|@{\hspace{1mm}}c@{\hspace{1mm}}|@{\hspace{1mm}}c@{\hspace{1mm}}|@{\hspace{1mm}}c@{\hspace{1mm}}|@{\hspace{1mm}}c@{\hspace{1mm}}} \hline
			& \multirow{2}*{$overall\_accu$} & \multicolumn{8}{c}{$class\_accu$} \\ \cline{3-10} 
			& & Wall & Door \& Window & Closet & Bathroom \& \etc & Living room \& \etc & Bedroom & Hall & Balcony \\ \hline \hline
			Raster-to-Vector~\cite{Raster2Vector} & 0.84 & 0.53 & 0.58 & 0.78 & 0.83 & 0.72 & \textbf{0.89} & 0.64 & 0.71 \\  \hline
			Ours & 0.88 & \textbf{0.88} & \textbf{0.86} & 0.80 & 0.86 & 0.86 & 0.75 & 0.73 & 0.86 \\ \hline
			Ours\dag & \textbf{0.89} & \textbf{0.88} & \textbf{0.86} & \textbf{0.82} & \textbf{0.90} & \textbf{0.87} & 0.77 & \textbf{0.82} & \textbf{0.93} \\ \hline
	\end{tabular}}
\vspace*{-2mm}
\end{table*}


\subsection{Network Training}
\label{ssec:network_training}


\paragraph{Datasets.}
As there are no public datasets with pixel-wise labels for floor plan recognition, we prepared two datasets, namely R2V and R3D.
Specifically, R2V has 815 images, all from Raster-to-Vector~\cite{Raster2Vector}, where the floor plans are mostly in rectangular shapes with uniform wall thickness.
For R3D, besides the original 214 images from~\cite{Rent3D}, we further added 18 floor plan images of round-shaped layouts to the data.
Compared with R2V, most room shapes in R3D are irregular with nonuniform wall thickness.
Here, we used Photoshop to manually label the image regions in R2V and R3D for walls, doors, bedrooms, etc. 
Note that we used the same label for some room regions,~\eg, living room and dining room (see Figure~\ref{fig3_1}), since they usually locate just next to one another without walls separating them.
Such a situation can be observed in both datasets.
Second, we followed the GitHub code in Raster-to-Vector~\cite{Raster2Vector} to group room regions, so that we can compare with their results.


For the train-test split ratio, we followed the original paper~\cite{Raster2Vector} to split R2V into 715 images for training and 100 images for testing.
For R3D, we randomly split it into 179 images for training and 53 images for testing.

%


\vspace*{-3mm}
\paragraph{Cross-and-within-task weighted loss.}
Each of the two tasks in our network involves multiple labels for various room-boundary and room-type elements.
Since the number of pixels varies for different elements, we have to balance their contributions within each task.
Also, there are generally more room-type pixels than room-boundary pixels, so we have to further balance the contributions of the two tasks.
Therefore, we design a {\em cross-and-within-task weighted loss} to balance between the two tasks as well as among the floor plan elements within each task.
\begin{itemize}
\item
{\em Within-task weighted loss\/}.
Here, we define the within-task weighted loss in an entropy style as
\vspace*{-2mm}
\begin{equation}
\label{eq:within_task}
\mathcal{L}_{task} = w_i \sum_{i=1}^C  -y_i \log{p_i},
\end{equation}
where
$y_i$ is the label of the $i$-th floor plan element in the floor plan and $C$ is the number of floor plan elements in the task;
$p_i$ is the prediction label of the pixels for the $i$-th element ($p_i \in [0,1]$); and
$w_i$ is defined as follows:
\vspace*{-2mm}
\begin{equation}
\label{eq:within_task_weight}
w_i
= \frac{\hat{\mathcal{N}} - {\hat{N}_i}}{\sum_{j=1}^C( \hat{\mathcal{N}} - {\hat{N}_j} )}, 
\end{equation}
where $\hat{N}_i$ is the total number of ground-truth pixels for the $i$-th floor plan element in the floor plan, and $\hat{\mathcal{N}}=\sum_{i=1}^C \hat{N}_i$, which means the total number of ground-truth pixels over all the $C$ floor plan elements.
\item
{\em Cross-and-within-task weighted loss\/}:
$L_{rb}$ and $L_{rt}$ denotes the within-task weighted losses for the room-boundary and room-type prediction tasks computed from Eq.~\eqref{eq:within_task}, respectively. $N_{rb}$ and $N_{rt}$ are the total number of network output pixels for room boundary and room type, respectively.
Then, the overall cross-and-within-task weighted loss $\mathcal{L}$ is defined as:
\vspace*{-2mm}
\begin{equation}
\label{eq:cross_task}
\mathcal{L} = w_{rb} \mathcal{L}_{rb} + w_{rt} \mathcal{L}_{rt},
\vspace*{-2mm}
\end{equation}
where $w_{rb}$ and $w_{rb}$ are weights given by
\vspace*{-2mm}
\begin{equation}
\label{eq:cross_task_weight}
w_{rb} = \frac{N_{rt}}{N_{rb}+N_{rt}} \ \ \text{and} \ \
w_{rt} = \frac{N_{rb}}{N_{rb}+N_{rt}}.
\end{equation}
\end{itemize}

\section{Experiments}

\begin{figure*}[t]
	\centering
	\includegraphics[width=0.98\linewidth]{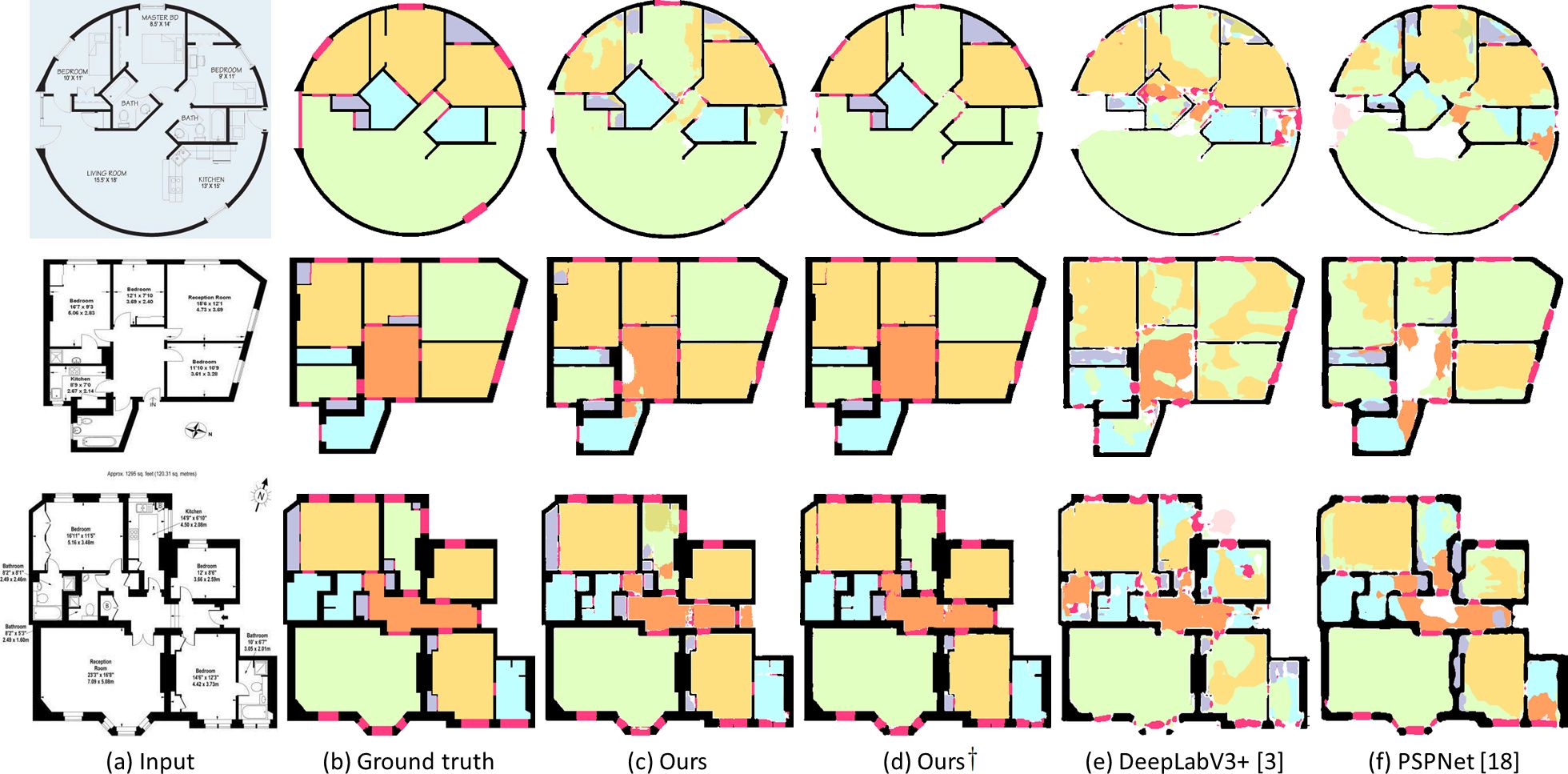}
	\caption{Visual comparison of floor plan recognition results produced by our method (c\&d) and  others (e-f) on the R3D dataset.
Symbol $\dag$ indicates our method with postprocessing (see Section~\ref{ssec:implem}).}
	\label{fig4_2}
\vspace*{-2mm}
\end{figure*}

\subsection{Implementation Details}
\label{ssec:implem}


\paragraph{Network training.}
We trained our network on an NVIDIA TITAN Xp GPU and ran 40k iterations in total.
We employed Adam optimizer to update the parameters and used a fixed learning rate of 1e-4 to train the network.
The resolution of the input floor plan is $512 \times 512$, for keeping the thin and short lines (such as the walls) in the floor plans.
Moreover, we used a batch size of one without using batch normalization, since it requires at least 32 batch size~\cite{Wu_2018_ECCV}.
Also, we did not use any other normalization method.
For other existing methods in our comparison, we used the original hyper-parameters reported in their original papers to train their networks. To obtain the best recognition results, we further evaluated the result every five training epochs and reported only the best one.



\vspace*{-4mm}
\paragraph{Network testing.}
\label{sec:post_process}
Given a test floor plan image, we feed it to our network and obtain its output.
However, due to the per-pixel prediction, the output may contain certain noise, so we further find connected regions bounded by the predicted room-boundary pixels to locate room regions, count the number of pixels of each predicted room type in each bounded region, and set the overall predicted type as the type of the largest frequency (see Figure~\ref{fig4_1}(c) \& (d)).
Our code and datasets are available at: {\small{\url{https://github.com/zlzeng/DeepFloorplan}}}.



\subsection{Qualitative and Quantitative Comparisons}

\paragraph{Comparing with Raster-to-Vector.}
First, we compared our method with Raster-to-Vector~\cite{Raster2Vector}, the state-of-the-art method for floor plan recognition.
Specifically, we used images from the R2V dataset to train its network and also our network.
To run Raster-to-Vector, we used its original labels (which are 2D corner coordinates of rectangular bounding boxes), while for our network, we used per-pixel labels.
Considering that the Raster-to-Vector network can only output 2D corner coordinates of bounding boxes, we followed the procedure presented in~\cite{Raster2Vector} to convert its bounding box outputs to per-pixel labels to facilitate comparison with our method; please refer to~\cite{Raster2Vector} for the procedural details.

Figure~\ref{fig4_1} (c-e) shows visual comparisons between our method and Raster-to-Vector.
For our method, we provide both results with (denoted with $\dag$) and w/o postprocessing.
For Raster-to-Vector, it has already contained a simple postprocessing step to connect room regions.
Comparing the results with the ground truths in (b), we can see that Raster-to-Vector tends to have poorer performance on room-boundary predictions,~\eg, missing even some room regions.
Our results are more similar to the ground truths, even without postprocessing.
For the R3D dataset, it contains many nonrectangular room shapes, so Raster-to-Vector performed badly with many missing regions, due to its Manhattan assumption; thus, we did not report the comparisons on R3D.

For quantitative evaluation, we adopted two widely-used metrics~\cite{Long_2015_CVPR},~\ie, the overall pixel accuracy and the per-class pixel accuracy:
\vspace*{-2mm}
\begin{equation}\label{eq4_accu}
overall\_accu = \frac{\sum_{i}N_{i}}{\sum_{i}\hat{N}_i} 
\ \ \text{and} \ \
class\_accu(i) = \frac{N_i}{\hat{N}_i} \ ,
\vspace*{-2mm}
\end{equation}
where $\hat{N}_i$ and $N_i$ are the total number of the ground-truth pixels and the correctly-predicted pixels for the $i$-th floor plan element, respectively.
Table~\ref{table4_5} shows the quantitative comparison results on the R2V dataset.
%
From the results, we can see that our method achieves higher accuracies for most floor plan elements, and the postprocessing could further improve our performance.

\begin{table*}[t]
	\centering
	\caption{Comparison with DeepLabV3+ and PSPNet.
	Besides the class accuracy, we further followed the GitHub code of~\cite{Long_2015_CVPR} to compute the $mean\_IoU$ metric; see the last row.
	The values inside () indicate the performance after postprocessing.
Note that the R2V dataset contains floor plans that are mostly in rectangular shapes, while the R3D dataset contains a much richer variety shape of floor plans.}
	\vspace*{1mm}
	\label{table4_3}
	\resizebox{0.95\linewidth}{!}{%
		\begin{tabular}{c||c||ll|ll|ll||ll|ll|ll} \hline
			& & \multicolumn{6}{c||}{R3D} & \multicolumn{6}{c}{R2V} \\ \cline{3-14}
			& & \multicolumn{2}{c|}{\small Ours} & \multicolumn{2}{c|}{\small DeepLabV3+~\cite{Chen_2018_ECCV}} & \multicolumn{2}{c||}{\small PSPNet~\cite{Zhao_2017_CVPR}} & \multicolumn{2}{c|}{\small Ours} & \multicolumn{2}{c|}{\small DeepLabV3+~\cite{Chen_2018_ECCV}} & \multicolumn{2}{c}{\small PSPNet~\cite{Zhao_2017_CVPR}} \\ \cline{2-14} \cline{1-14}
			$overall\_accu$ & & 0.89 & (\textbf{0.90}) & 0.85 & (0.83) & 0.84 & (0.81) & 0.89 & (\textbf{0.90})  & 0.88 & (0.87) & 0.88 & (0.88) \\ \hline
			\multirow{8}{*}{$class\_accu$} & wall   & 0.98 & (\textbf{0.98})  & 0.93 &(0.93) & 0.91 &(0.91) & 0.89 &(\textbf{0.89})  & 0.80 &(0.80) & 0.84 &(0.84) \\ \cline{2-14}
			& door-and-window & 0.83 &(\textbf{0.83}) & 0.60 &(0.60) & 0.54& (0.54) & 0.89 &(\textbf{0.89}) & 0.72& (0.72) & 0.76 &(0.76) \\ \cline{2-14}
			& closet       & \textbf{0.61} &(0.54) & 0.24 &(0.048) & 0.45 &(0.086) & 0.81 &(\textbf{0.92}) & 0.78& (0.85) & 0.80 &(0.71) \\ \cline{2-14}
			& bathroom \& \etc     & \textbf{0.81}& (0.78) & 0.76& (0.57) & 0.70& (0.50) & 0.87& (\textbf{0.93}) & 0.90& (0.90) & 0.90 &(0.84) \\ \cline{2-14}
			& living room \& \etc   & 0.87& (\textbf{0.93}) & 0.76 &(0.90) & 0.76& (0.89) & 0.88& (\textbf{0.91}) & 0.85& (0.84) & 0.83 &(0.90) \\ \cline{2-14}
			& bedroom      & 0.75& (\textbf{0.79}) & 0.56& (0.40) & 0.55 &(0.40) & 0.83& (\textbf{0.91}) & 0.82& (0.65) & 0.86& (0.92) \\ \cline{2-14}
			& hall         & 0.59 &(0.68) & \textbf{0.72}& (0.44) & 0.61 &(0.23) & 0.68 &(0.84) & 0.55& (\textbf{0.87}) & 0.78 &(0.81) \\ \cline{2-14}
			& balcony      & 0.44 &(\textbf{0.49}) & 0.08& (0.0027) & 0.41 &(0.11) & 0.90& (\textbf{0.92}) & 0.87& (0.45) & 0.87 &(0.82) \\ \hline
			$mean\_IoU$ &  & 0.63 & (\textbf{0.66}) & 0.50 & (0.44) & 0.50 & (0.41) & 0.74 & (\textbf{0.76})  & 0.69 & (0.67) & 0.70 & (0.69) \\ \hline			
	\end{tabular}}
	\vspace*{-2mm}
\end{table*}

\vspace*{-4mm}
\paragraph{Comparing with segmentation networks.}
To evaluate how general segmentation networks perform for floor plan recognition, we further compare our method with two recent segmentation networks, DeepLabV3+~\cite{Chen_2018_ECCV} and PSPNet~\cite{Zhao_2017_CVPR}.

For a fair comparison, we trained their networks, as well as our network, on the R2V dataset and also on the R3D dataset, and adjusted their hyper-parameters to obtain the best recognition results.
Figures~\ref{fig4_1} \&~\ref{fig4_2} present visual comparisons with PSPNet and DeepLabV3+ on testing floor plans from R2V and R3D, respectively.
Due to space limitation, please see our supplementary material for results of PSPNet and DeepLabV3+ with postprocessing.
From the figures, we can see that their results tend to contain noise, especially for complex room layouts and small elements like doors and windows.
Since these elements are usually the room boundary between room regions, so the results further affect the room-type predictions.
Please see the supplementary material for more visual comparison results.

Table~\ref{table4_3} reports the quantitative comparison results for various methods with and without postprocessing, in terms of the overall and per-class accuracy, on both R2V and R3D datasets.
Comparing with DeepLabV3+ and PSPNet, our method performs better for most floor plan elements, even without postprocessing, showing its superiority over these general-purpose segmentation networks.
Note that, our postprocessing step assumes plausible room-boundary predictions, so it typically fails to enhance results with poor room-boundary predictions; see the results in Figure~\ref{fig4_2}.


\paragraph{Comparing with an edge detection method.} \
To show that room boundaries (\ie, wall, door, and window) are not merely edges in the floor plans but structural elements with semantics, we further compare our method with a state-of-the-art edge detection network~\cite{liu2017richer} (denoted as RCF) on detecting wall elements in floor plans.
Here, we re-trained RCF using our wall labels, separately on the R2V and R3D datasets; since RCF outputs a per-pixel probability ($\in [0,1]$) on wall prediction, we need a threshold (denoted as $t_\text{RCF}$) to locate the wall pixels from its results.
In our method, we extract a binary map from our network output for walls pixels; see Figure~\ref{fig3_1} (bottom) for an example.

To quantitatively compare the binary maps produced by RCF and our method, we employ F-measure~\cite{Hou_2018}, a commonly-used metric, which is expressed as
\vspace*{-1mm}
\begin{equation}\label{eq4_f_beta}
F_{\beta} = \frac{(1+\beta^2)Precision \times Recall}{\beta^2Precision + Recall} \ ,
\vspace*{-1mm}
\end{equation}
where $Precision$ and $Recall$ are the ratios of the correctly-predicted wall pixels over all the predicted wall pixels and over all the ground-truth wall pixels, respectively.
To account for the fact that we need $t_\text{RCF}$ to threshold RCF's results, we extend $F_{\beta}$ into $F_{\beta}^\text{max}$ and $F_{\beta}^\text{mean}$ in the evaluations:
\vspace*{-1.5mm}
\begin{displaymath}
F_{\beta}^\text{max} =
\frac{1}{M}\sum_{p=1}^{M}{\tilde{F}^p_{\beta}}
\ \ \text{and} \ \
F_{\beta}^\text{mean} =
\frac{1}{MT}\sum_{p=1}^{M}\sum_{t=0}^{T-1}{F^p_{\beta}({\small \frac{t}{T-1}})},
\vspace*{-1mm}
\end{displaymath}
%
%
where $M$ is the total number of testing floor plans;
$\tilde{F}^p_\beta$ is the best $F_{\beta}$ on the $p$-th test input over $T$ different $t_\text{RCF}$ ranged in [0,1];
and
$F^p_{\beta}({\small \frac{t}{T-1}})$ is $F_{\beta}$ on the $p$-th test input using $t_\text{RCF}=\frac{t}{T-1}$.
In our implementation, as suggested by previous work~\cite{Hou_2018}, we empirically set $\beta^2$$=$$0.3$ and $T$$=$$256$.
Note that $F_{\beta}^\text{max}$ and $F_{\beta}^\text{mean}$ are the same for the binary maps produced by our method, since they do not require $t_\text{RCF}$.
Table~\ref{table4_2} reports the results, clearly showing that our method outperforms RCF on detecting the walls.
Having said that, simply detecting edges in the floor plan images is inefficient to floor plan recognition.


\begin{table}[!t]
	\centering
	\caption{Comparison with a state-of-the-art edge detection network (RCF~\cite{liu2017richer}) on detecting the walls in floor plans.}
	\label{table4_2}
	\vspace*{1mm}
	\resizebox{0.68\linewidth}{!}{
		\begin{tabular}{c||c|c||c|c} \hline
			& \multicolumn{2}{c||}{R2V} & \multicolumn{2}{c}{R3D} \\ \cline{2-5}
			& $F_{\beta}^\text{max}$ & $F_{\beta}^\text{mean}$ & $F_{\beta}^\text{max}$ & $F_{\beta}^\text{mean}$ \\ \hline \hline
			RCF~\cite{liu2017richer} & 0.62 & 0.56 & 0.68 & 0.58 \\ \hline
			Ours & \textbf{0.85} & \textbf{0.85} & \textbf{0.95} & \textbf{0.95} \\ \hline
	\end{tabular}}
\end{table}

\begin{figure*}[!ht]
	\centering
	\includegraphics[width=0.89\linewidth]{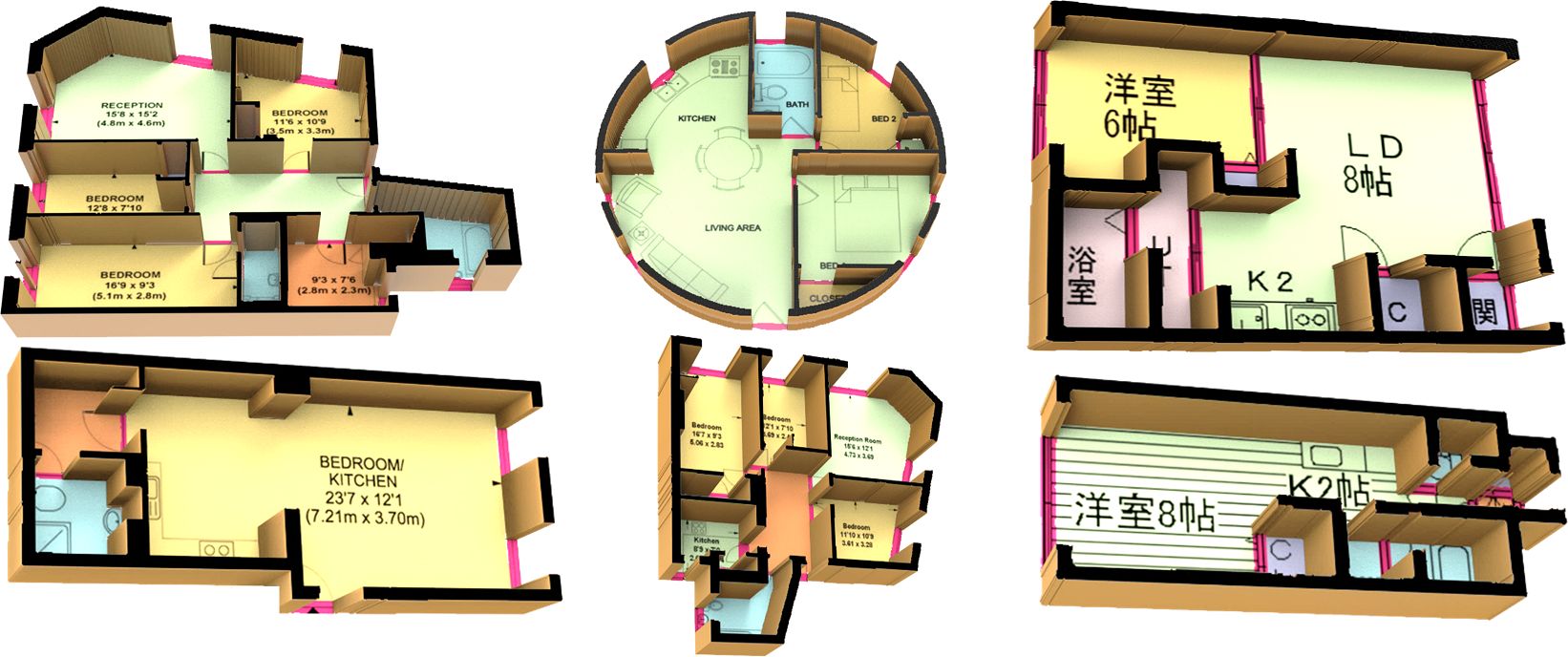}
	\caption{Reconstructed 3D models from our recognition results.}
	\label{fig6_1}
	\vspace*{-2mm}
\end{figure*}

\subsection{Architecture Analysis on our Network}
Next, we present an architecture analysis on our network by comparing it with the following two baseline networks:
\begin{itemize}
	\vspace*{-1mm}
	\item
	{\em Baseline \#1: two separate single-task networks.}
	The first baseline breaks the problem into two separate single-task networks, one for room-boundary prediction and the other for room-type prediction, with two separate sets of VGG encoders and decoders.
Hence, there are no shared features and also no spatial contextual modules compared to our full network.
	\vspace*{-1mm}
	\item
	{\em Baseline \#2: without the spatial contextual module.}
	The second baseline is our full network with the shared features but without the spatial contextual module.
\end{itemize}

\begin{table}[!t]
	\centering
	\caption{A comparison of our full network with {\em Baseline network \#1} and {\em Baseline network \#2} using the R3D dataset.}
	\vspace*{1mm}
	\label{tab:multitask}
	\resizebox{1.0\linewidth}{!}{
		\begin{tabular}{c||c|c|c} \hline
			\multirow{2}*{Metrics}	 & \multicolumn{3}{c}{Methods} \\ \cline{2-4}
			& {\small Baseline \#1} & {\small Baseline \#2} & {\small Our full network} \\ \hline \hline
			{$overall\_accu$} & 0.82                  & 0.85                  & \textbf{0.89} \\ \hline
			{$average \ class\_accu$}    & 0.72                  & 0.72                  & \textbf{0.80} \\ \hline
		\end{tabular}
	}
	\vspace*{-2mm}
\end{table}

Table~\ref{tab:multitask} shows the comparison results, where we trained and tested each network using the R3D dataset~\cite{Rent3D}. 
From the results, we can see that our full network outperforms the two baselines, indicating that the multi-task scheme with the shared features and the spatial contextual module both help improve the floor plan recognition performance.


\subsection{Analysis on the Spatial Contextual Module}
An ablation analysis of the spatial contextual module (see Figure~\ref{fig3_4} for details) is presented here. 
\begin{itemize}
	%
	\item
	{\em No attention}: the room-boundary-guided attention mechanism (see the top branch in Figure~\ref{fig3_4}) is removed from the spatial contextual module.
	\vspace*{-2mm}
	\item
	{\em No direction-aware kernels}: the convolution layers with the four direction-aware kernels in the spatial contextual module are removed. Only the room-boundary-guided attention mechanism is applied.
\end{itemize}

Table~\ref{tab:ablation} shows the comparison results between the above schemes and the full method (\ie, with both attention and direction-aware kernels).
Again, we trained and tested on the R3D dataset~\cite{Rent3D}.
From Table~\ref{tab:ablation}, we can see that the spatial contextual module performs the best when equipped with the attention mechanism and direction-aware kernels.

\subsection{Discussion}

\paragraph{Application: 3D model reconstruction.}
Here, we take our floor plan recognition results to reconstruct 3D models.
Figure~\ref{fig6_1} shows several examples of the constructed 3D floor plans. Our method is able to recognize walls of nonuniform thickness and a wide variety of shapes. It thus enables us to construct 3D room-boundary of various shapes,~\eg, curved walls in floor plan.
One may notice that we only reconstruct the walls in 3D in Figure~\ref{fig6_1}. In fact, we may further reconstruct the doors and windows, since our method has also recognized them in the layouts. For more reconstruction results, please refer to our supplementary material.

\begin{table}[t]
	\centering
	\caption{Ablation study on the spatial contextual module.}
	\vspace*{1mm}
	\label{tab:ablation}
	\resizebox{0.9\linewidth}{!}{
		\begin{tabular}{c||@{\hspace{1mm}}c@{\hspace{1mm}}|@{\hspace{1mm}}c@{\hspace{1mm}}|@{\hspace{1mm}}c@{\hspace{1mm}}} \hline
			\multirow{3}*{Metrics} & \multicolumn{3}{c}{Methods} \\ \cline{2-4}
			& {\multirow{2}*{\small No attention}} & {\small No direction} & {\small Our complete} \\
			&                                      & {\small -aware kernels}            & {\small version}      \\ \hline \hline
			{$overall\_accu$}      & 0.86                                & 0.87                        & \textbf{0.89} \\ \hline
			{$average \ class\_accu$}& 0.74                               & 0.77                        & \textbf{0.80} \\ \hline
	\end{tabular}}
	\vspace*{-2mm}
\end{table}

\vspace*{-4mm}
\paragraph{Limitations.}
Here, we discuss two challenging situations, for which our method fails to produce plausible predictions.
First, our network may fail to differentiate inside and outside regions, in case there are some special room structures in the floor plan,~\eg, long and double-bended corridors.
Second, our network may wrongly recognize large icons (\eg, compass icon) in floor plans as wall elements.
To address these issues, we believe that more data is needed for the network to learn more variety of floor plans and the semantics.
Also, we may explore weakly-supervised learning for the problem to avoid the tedious annotations; please see the supplemental material for example failure cases.

\section{Conclusion}

This paper presents a new method for recognizing floor plan elements.
There are three key contributions in this work.
First, we explore the spatial relationship between floor plan elements, model a hierarchy of floor plan elements, and design a multi-task network to learn to recognize room-boundary and room-type elements in floor plans.
Second, we further take the room-boundary features to guide the room-type prediction by formulating the spatial contextual module with the room-boundary-guided attention mechanism.
%
Further, we design a cross-and-within-task weighted loss to balance the losses within each task and across tasks.
In the end, we prepared also two datasets for floor plan recognition and extensively evaluated our network in various aspects.
Results show the superiority of our network over the others in terms of the overall accuracy and $F_{\beta}$ metrics.
In the future, we plan to further extract the dimension information in the floor plan images, and learn to recognize the text labels and symbols in floor plans.


{\small
\bibliographystyle{ieee_fullname}
\bibliography{floorlayout}

\begin{thebibliography}{10}\itemsep=-1pt

\bibitem{ah1997variations}
Christian Ah-Soon and Karl Tombre.
\newblock {V}ariations on the analysis of architectural drawings.
\newblock In {\em International Conference on Document Analysis and Recognition
  (ICDAR)}. IEEE, 1997.

\bibitem{ahmed2011improved}
Sheraz Ahmed, Marcus Liwicki, Markus Weber, and Andreas Dengel.
\newblock {I}mproved automatic analysis of architectural floor plans.
\newblock In {\em International Conference on Document Analysis and Recognition
  (ICDAR)}. IEEE, 2011.

\bibitem{Chen_2018_ECCV}
Liang-Chieh Chen, Yukun Zhu, George Papandreou, Florian Schroff, and Hartwig
  Adam.
\newblock {E}ncoder-decoder with atrous separable convolution for semantic
  image segmentation.
\newblock In {\em European Conference on Computer Vision (ECCV)}, 2018.

\bibitem{de2011wall}
Llu{\'\i}s-Pere de~las Heras, Joan Mas, Gemma Sanchez, and Ernest Valveny.
\newblock {W}all patch-based segmentation in architectural floorplans.
\newblock In {\em International Conference on Document Analysis and Recognition
  (ICDAR)}. IEEE, 2011.

\bibitem{dodge2017parsing}
Samuel Dodge, Jiu Xu, and Bj{\"o}rn Stenger.
\newblock {P}arsing floor plan images.
\newblock In {\em International Conference on Machine Vision Applications
  (MVA)}. IEEE, 2017.

\bibitem{dosch2000complete}
Philippe Dosch, Karl Tombre, Christian Ah-Soon, and G{\'e}rald Masini.
\newblock {A} complete system for the analysis of architectural drawings.
\newblock {\em International Journal on Document Analysis and Recognition},
  3(2):102--116, 2000.

\bibitem{gimenez2016automatic}
Lucile Gimenez, Sylvain Robert, Fr{\'e}d{\'e}ric Suard, and Khaldoun Zreik.
\newblock {A}utomatic reconstruction of {3D} building models from scanned {2D}
  floor plans.
\newblock {\em Automation in Construction}, 63:48--56, 2016.

\bibitem{Hou_2018}
Qibin Hou, Ming-Ming Cheng, Xiaowei Hu, Ali Borji, Zhuowen Tu, and Philip H.~S.
  Torr.
\newblock {D}eeply supervised salient object detection with short connections.
\newblock {\em IEEE Transactions on Pattern Analysis and Machine Intelligence},
  41(4):815--828, 2018.

\bibitem{Lee_2017_ICCV}
Chen-Yu Lee, Vijay Badrinarayanan, Tomasz Malisiewicz, and Andrew Rabinovich.
\newblock {R}oom{N}et: {E}nd-to-end room layout estimation.
\newblock In {\em IEEE International Conference on Computer Vision (ICCV)},
  2017.

\bibitem{Rent3D}
Chenxi Liu, Alex Schwing, Kaustav Kundu, Raquel Urtasun, and Sanja Fidler.
\newblock {R}ent3{D}: {F}loor-plan priors for monocular layout estimation.
\newblock In {\em IEEE Conference on Computer Vision and Pattern Recognition
  (CVPR)}, 2015.

\bibitem{Raster2Vector}
Chen Liu, Jiajun Wu, Pushmeet Kohli, and Yasutaka Furukawa.
\newblock {R}aster-to-{V}ector: {R}evisiting floorplan transformation.
\newblock In {\em IEEE International Conference on Computer Vision (ICCV)},
  2017.

\bibitem{liu2017richer}
Yun Liu, Ming-Ming Cheng, Xiaowei Hu, Kai Wang, and Xiang Bai.
\newblock {R}icher convolutional features for edge detection.
\newblock In {\em IEEE Conference on Computer Vision and Pattern Recognition
  (CVPR)}, 2017.

\bibitem{Long_2015_CVPR}
Jonathan Long, Evan Shelhamer, and Trevor Darrell.
\newblock {F}ully convolutional networks for semantic segmentation.
\newblock In {\em IEEE Conference on Computer Vision and Pattern Recognition
  (CVPR)}, 2015.

\bibitem{mace2010system}
S{\'e}bastien Mac{\'e}, Herv{\'e} Locteau, Ernest Valveny, and Salvatore
  Tabbone.
\newblock {A} system to detect rooms in architectural floor plan images.
\newblock In {\em Proceedings of the 9th IAPR International Workshop on
  Document Analysis Systems}, 2010.

\bibitem{or2005highly}
Siu-Hang Or, Kin-Hong Wong, Ying-Kin Yu, and Michael Ming-Yuan Chang.
\newblock {H}ighly automatic approach to architectural floorplan image
  understanding and model generation.
\newblock In {\em Proc. of Vision, Modeling, and Visualization 2005
  (VMV-2005)}, pages 25--32, 2005.

\bibitem{ryall1995semi}
Kathy Ryall, Stuart Shieber, Joe Marks, and Murray Mazer.
\newblock {S}emi-automatic delineation of regions in floor plans.
\newblock In {\em International Conference on Document Analysis and Recognition
  (ICDAR)}. IEEE, 1995.

\bibitem{Simonyan14c}
Karen Simonyan and Andrew Zisserman.
\newblock {V}ery deep convolutional networks for large-scale image recognition.
\newblock In {\em International Conference on Learning Representations (ICLR)},
  2015.

\bibitem{Sun_2019_CVPR}
Cheng Sun, Chi-Wei Hsiao, Min Sun, and Hwann-Tzong Chen.
\newblock {H}orizon{N}et: Learning room layout with 1{D} representation and
  pano stretch data augmentation.
\newblock In {\em IEEE Conference on Computer Vision and Pattern Recognition
  (CVPR)}, 2019.

\bibitem{Wu_2018_ECCV}
Yuxin Wu and Kaiming He.
\newblock {G}roup normalization.
\newblock In {\em European Conference on Computer Vision (ECCV)}, 2018.

\bibitem{yamasaki2018apartment}
Toshihiko Yamasaki, Jin Zhang, and Yuki Takada.
\newblock {A}partment structure estimation using fully convolutional networks
  and graph model.
\newblock In {\em Proceedings of the 2018 ACM Workshop on Multimedia for Real
  Estate Tech}, 2018.

\bibitem{Yang_2019_CVPR}
Shang-Ta Yang, Fu-En Wang, Chi-Han Peng, Peter Wonka, Min Sun, and Hung-Kuo
  Chu.
\newblock {D}u{L}a-{N}et: A dual-projection network for estimating room layouts
  from a single {RGB} panorama.
\newblock In {\em IEEE Conference on Computer Vision and Pattern Recognition
  (CVPR)}, 2019.

\bibitem{Zhang_2014_ECCV}
Yinda Zhang, Shuran Song, Ping Tan, and Jianxiong Xiao.
\newblock {P}ano{C}ontext: {A} whole-room 3{D} context model for panoramic
  scene understanding.
\newblock In {\em European Conference on Computer Vision (ECCV)}, 2014.

\bibitem{Zhao_2017_CVPR}
Hengshuang Zhao, Jianping Shi, Xiaojuan Qi, Xiaogang Wang, and Jiaya Jia.
\newblock {P}yramid scene parsing network.
\newblock In {\em IEEE Conference on Computer Vision and Pattern Recognition
  (CVPR)}, 2017.

\bibitem{Zou_2018_CVPR}
Chuhang Zou, Alex Colburn, Qi Shan, and Derek Hoiem.
\newblock {L}ayout{N}et: {R}econstructing the 3{D} room layout from a single
  {RGB} image.
\newblock In {\em IEEE Conference on Computer Vision and Pattern Recognition
  (CVPR)}, 2018.

\end{thebibliography}
}

\end{document}